\documentclass[letterpaper, 10 pt, conference]{ieeeconf}  % Comment this line out if you need a4paper

\usepackage{multicol}
\usepackage[bookmarks=true]{hyperref}
\usepackage{amsmath}
\usepackage{amsfonts}
\usepackage{mathtools}
\usepackage{algorithm}
\usepackage[noend]{algpseudocode}
\usepackage{subcaption}
\usepackage{bm}
\usepackage{balance}
\usepackage{verbatim}

\title{Prediction of Human Full-Body Movements with\\ Motion Optimization and Recurrent Neural Networks}

\author{Philipp Kratzer$^{1}$, Marc Toussaint$^{1}$ and Jim Mainprice$^{1,2}$\\% <-this % stops a space
\vspace{0.1cm}
\authorblockA{\tt{\small{firstname.lastname@ipvs.uni-stuttgart.de}}}
\authorblockA{$^1$Machine Learning and Robotics Lab, University of Stuttgart, Germany}
\authorblockA{$^2$Max Planck Institute for Intelligent Systems ;  IS-MPI ; T{\"u}bingen, Germany}
\vspace{-0.8cm}
}

\begin{document}

\graphicspath{{.}{figures/}}

\maketitle
\thispagestyle{empty}
\pagestyle{empty}

%%%%%%%%%%%%%%%%%%%%%%%%%%%%%%%%%%%%%%%%%%%%%%%%%%%%%%%%%%%%%%%%%%%%%%%%%%%%%%%%
\begin{abstract}
Human movement prediction is difficult as humans naturally exhibit
complex behaviors that can change drastically from one environment to the next.
In order to alleviate this issue, we propose a prediction framework that
decouples short-term prediction,
linked to internal body dynamics, and long-term prediction,
linked to the environment and task constraints.
In this work we investigate encoding short-term dynamics
in a recurrent neural network, while
we account for environmental constraints, such as obstacle avoidance,
using gradient-based trajectory optimization.
Experiments on real motion data demonstrate that our framework improves the prediction with respect to state-of-the-art motion prediction
methods, as it accounts to beforehand unseen environmental structures.
Moreover we demonstrate on an example, how this framework
can be used to plan robot trajectories that are optimized
to coordinate with a human partner.
\end{abstract}

%%%%%%%%%%%%%%%%%%%%%%%%%%%%%%%%%%%%%%%%%%%%%%%%%%%%%%%%%%%%%%%%%%%%%%%%%%%%%%%%
\section{INTRODUCTION}
For safe and efficient human-robot interaction it is crucial to foresee human motion in order to plan around the human partner and interact with the human partner without disturbing the natural flow of the human's motion. A good interaction strategy needs to plan  trajectories that minimally intervene with the human while still retaining the ability that both, the human and the robot, can achieve their goals without having to deviate widely from their optimal trajectory.

However, human motion is the result of complex biomechanical processes that are challenging to model. As a consequence, state-of-the-art work on motion prediction focuses on data-driven models, such as recurrent neural network models~\cite{martinez2017human, pavllo2018quaternet, wang2019vred}. A drawback of these architectures is that the network is only trained on the human state and therefore not able to take scene context, such as targets for reaching motion or obstacles into account. Adapting scene context directly into the architecture would require a generalizable scene representation and huge amounts of training data to be able to generalize to unseen environments.

In prior work~\cite{kratzer2018}, we proposed to account for environmental constraints in a later trajectory optimization step, using a Gaussian Process (GP)
to model the low-level dynamics.
However, GPs do not scale to large datasets of training
data as they require comparing all training points
in the data set to predict the next state.
In this work, we instead propose to adapt a state-of-the-art recurrent neural network model~\cite{wang2019vred} to learn purely kinematic predictions of the human.
In order to optimize the human motion in a later stage,
we introduce a modification to the network architecture to control
the velocities of the human at each prediction step.
We can then differentiate the network with respect to this control input,
and optimize the motion using a gradient-based optimization algorithm.

\begin{figure}
\centering
\includegraphics[width=\linewidth]{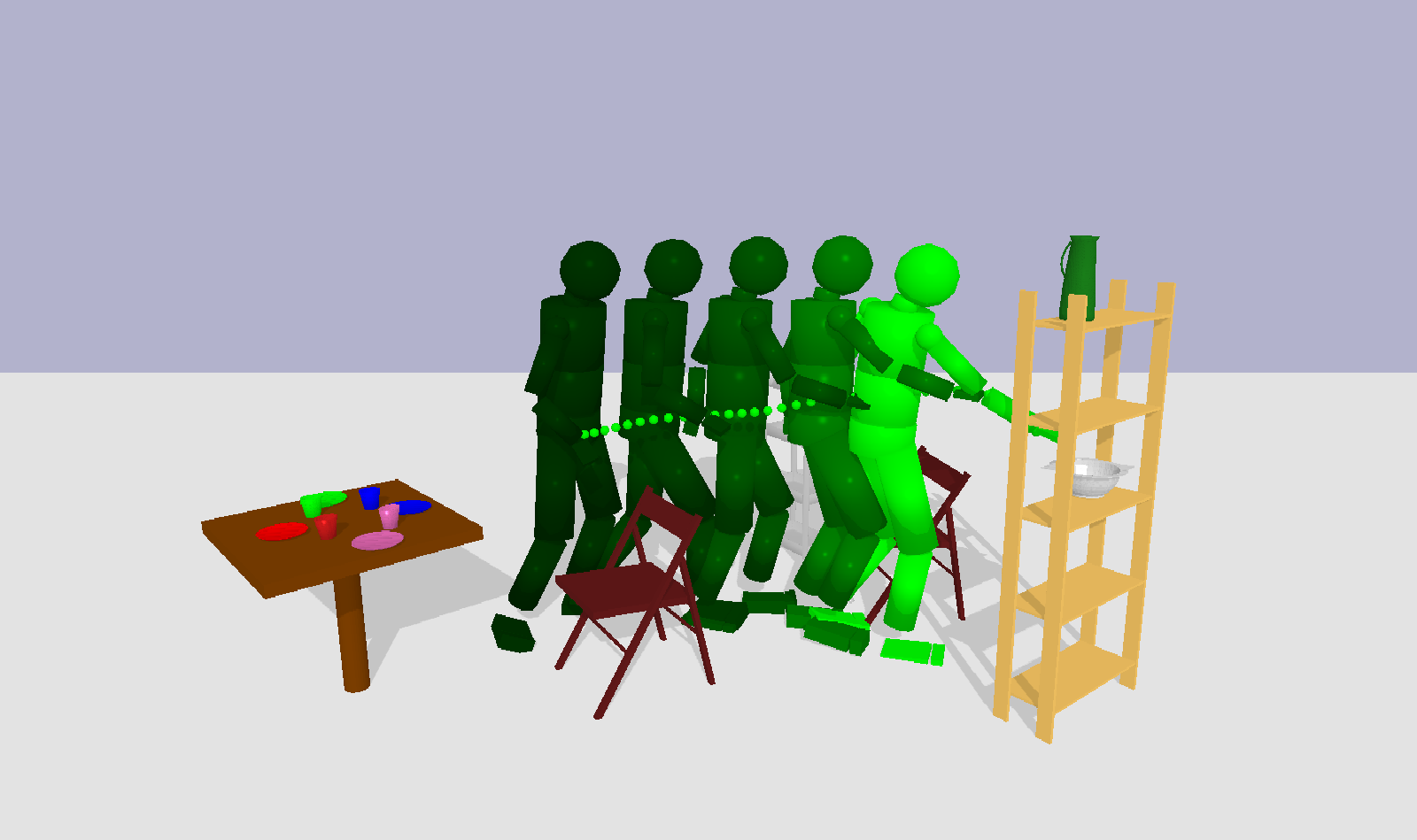}
\caption{Prediction of 1.5sec reaching motion towards the bowl on the big shelf by our method}
\label{fig:pred_plate}
\vspace{-.5cm}
\end{figure}

While the idea of blending motion capture data and motion planning
algorithms for motion prediction is not new \cite{pettre20032},
to the best of our knowledge,
this paper is the first to combine motion optimization with a recurrent neural network in order to predict human motion.
This approach has the following advantages: 1) It
relies on potentially infinite amount
of motion capture data to train the low-level dynamics as
the network does not grow with
the size of the training dataset,
2) Optimization is a flexible framework for
motion planning, which allows to integrate many
different constraints
(i.e.,  smoothness, obstacle avoidance,
closed kinematic chain, hand orientation, goalset, human-robot interaction, etc.).

In this work, we present experiments on motion capture data
recorded at the University of Stuttgart.
We test our approach on segmented motions ranging between 1 and 4 seconds
as shown in Figure \ref{fig:pred_plate}
using goalset constraints, obstacle avoidance and joint human-robot coordination.
Our results indicate that the additional motion optimization phase leads
to higher predictive performance than state-of-the-art neural network predictors,
especially in long-term prediction.

This paper is organized as follows: In Section~\ref{sec:related_work} we discuss relevant prior work. Section~\ref{sec:method} introduce our framework theoretically and explain how the implementation is done.
In Section~\ref{sec:experiments} we evaluate our prediction framework on real motion data. Conclusions are drawn in Section~\ref{sec:conclusions}.
\section{RELATED WORK}
\label{sec:related_work}
\subsection{Human Motion Prediction in Robotics}

Prior work on motion prediction in robotics has made use of graphical models. For example, Kuli{\'c} et al. encoded full-body motion primitives using Hidden Markov Models and applied the model to motion imitation~\cite{kulic2012incremental}. Koppula and Saxena focused on movement prediction using conditional random fields~\cite{koppula2016anticipating}.
While these approaches are sound they generally do not scale
to large databases of motion capture.

Another approach commonly used for predicting human motion is Inverse Optimal Control (IOC),
which aims to find a cost function underlying the observed behavior. For example, Berret et al. investigated cost functions for arm movement planning~\cite{berret2011evidence}. The authors reported that such movements are closely related to mechanical energy expenditure and joint-level smoothness. Mainprice et al. investigated prediction of human reaching motions in shared workspaces~\cite{mainprice2016goal}. Their method accounts for obstacles and a moving collaborator using iterative replanning. In IOC, bio-kinematic processes are typically represented by simplified models, which are not able to completely capture the complex bio-mechanical behaviors of human full-body motion and to accurately forecast an observed motion.

\subsection{Neural Network Human Motion Prediction}
Recent work on human motion prediction for short-term motion has focused on recurrent neural network architectures (RNN). Fragkiadaki et al. proposed a RNN based model that incorporates nonlinear encoder and decoder networks before and after recurrent layers~\cite{fragkiadaki2015recurrent}. Their model is able to handle training across multiple subjects and activity domains. Jain et al. introduced a method to incorporate structural elements into a RNN architecture~\cite{jain2016structural}. Autoencoders also can be used for denoising the prediction~\cite{ghosh2017learning}. Martinez et al. introduced a gated recurrent unit (GRU) based approach with a residual connection in the loop function and showed that this outperforms prior RNN based methods~\cite{martinez2017human}. Pavllo et al. further improved the RNN-based prediction by changing the joint angle representation to quaternions~\cite{pavllo2018quaternet, pavllo2019modeling}. However, this comes at the cost of additional normalization layers and normalization penalty. Recently Wang and Feng introduced a position-velocity recurrent encoder-decoder model (VRED)~\cite{wang2019vred}. Their model adds an additional velocity connection as an input to the GRU cell in the recurrent structure.

Motion prediction based on recurrent neural networks promises good results for predicting short-term motion. However, the models are trained on human data only. Handling environmental constraints is not possible yet and would require large amounts of training data.
Human environments are typically cluttered with objects and obstacles. Thus, in a human-robot collaboration scenario, adapting to such environmental constraints is crucial for the prediction.

For our human prediction model we adapt the VRED architecture by Wang and Feng and modify it as described in Subsection~\ref{ssec:human_model}.

\subsection{Motion Optimization}
Gradient-based optimization algorithms are widely used in the field of robotics and optimal control~\cite{todorov2005generalized, ratliff2009chomp, schulman2013finding, toussaint2014newton, marinho2016functional, Toussaint:17} for optimizing trajectories.
These techniques have been shown to successfully generate motions with
a variety of kinematic and dynamic objective and constraints.
In 2009 \cite{ratliff2009chomp},
Ratliff et al. have used insights from differential geometry
making it possible to use gradient-based optimization to
solve motion planning with non-convex obstacles.
Their main finding was that obstacle costs should be integrated
with respect to arc-length in Cartesian space,
leading to the notion of workspace geodesics ~\cite{ratliff2015understanding, Mainprice:16}, which we use here.

Mordatch et al. use motion optimization techniques to synthesize complex behaviors~\cite{mordatch2012discovery, mordatch2013animating}. The authors show that motion optimization approaches for synthesizing motion and animating characters can generate realistic motions even without the use of motion capture data. While the focus of their work does not lie on forecasting an observed motion, concepts, such as constraining foot contacts, have great potential to be incorporated in our motion prediction framework.
% TODO: motion synthesis literature (CG)

\section{METHOD}
\label{sec:method}

Next we formalize our approach, by presenting
first our method for predicting human motion and then
for planning a coordinated human-robot behavior.

\begin{figure*}
\center0
  \includegraphics[width=.7\linewidth]{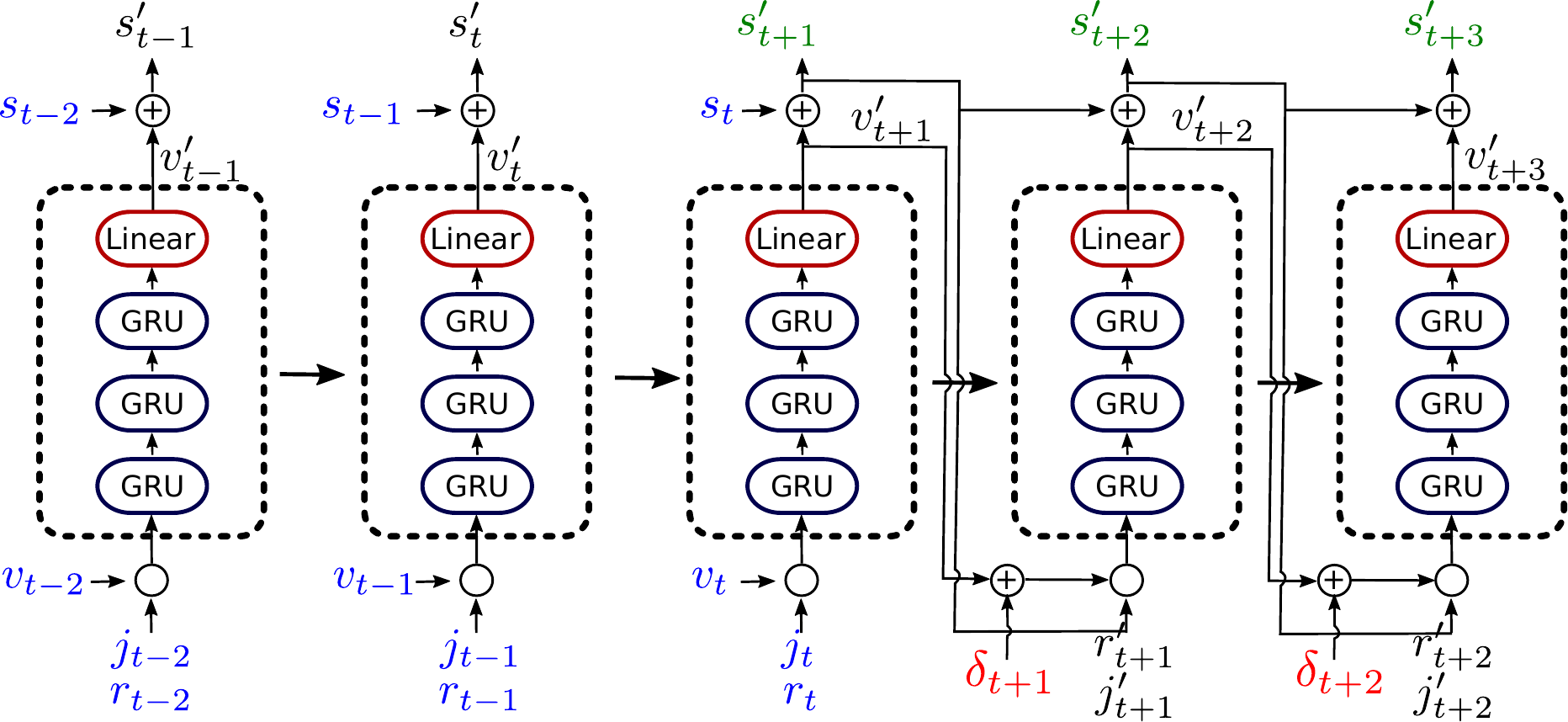}
  \caption{Architecture of the Recurrent Neural Network $s_{t+1:T} = f(s_{0:t})$. 
  In blue the past joint angles and velocities,  in green the future predicted states and in red the control input $\delta$, which allows motion optimization. }
  \label{fig:architecture}
  \vspace{-.5cm}
\end{figure*}

\subsection{Human Motion Optimization}
When predicting human motion,
the aim is to find a trajectory $s_{t+1:T}^*$ that maximizes the likelihood
of the next states given our demonstration data $\mathcal{D}$ and fulfills the constraints. The constraints are the human dynamics $s_{t+1} = d(s_t)$ and further constraints $h$ and $g$ that arise from the environment.

This can be written as the following optimization problem:
\begin{align}
  \min_{s_{t+1:T}}~\sum_{t+1}^T -\log p(s_{t+1}|s_{0:t}, \mathcal{D})&\\
    \text{subject to } s_{t+1} &= d(s_t) \notag\\
    h(s_{t}) &= 0 \notag\\
    g(s_{t}) &\leq 0 \notag
\end{align}

We approximate the maximum likelihood given $\mathcal{D}$ as a regression problem and train a recurrent neural network model to find the prediction states $s_{t+1:T} = f(s_{0:t}, \delta_{t+1:T})$ as discussed in Subsection~\ref{ssec:human_model}. We introduce parameters $\delta_{t+1:T}$ to the model in order to be able to vary the predictions and thus account for the constraints $h$ and $g$ using numerical optimization. We will describe this in Subsection~\ref{ssec:trajopt}.
The dynamic constraint $s_{t+1} = d(s_t)$ is learned from the data and thus also approximated by the recurrent neural network~$f$.

\subsection{Coordinated Motion Optimization}
In the case of joint motion optimization of the human and the robot, we not only aim to find a human trajectory $s_{t+1:T}^*$ but also a robot trajectory $x_{t+1:T}^*$. The optimization problem changes to:
\begin{align}
  \min_{s_{t+1:T},x_{t+1:T}}~\sum_{t+1}^T -\log p(s_{t+1} | s_{0:t}, \mathcal{D})& + c_R(x_{t+1:T})\\
    \text{subject to } s_{t+1}, x_{t+1} &= d(s_t, x_t) \notag\\
    h(s_t, x_t) &= 0 \notag\\
    g(s_t, x_t) &\leq 0 \notag
\end{align}
with $C_R$ being costs on the robot trajectory $x_{t+1:T}$.
The additional constraints $h$ and $g$ can now be function of both, $s_t$ and $x_t$, and arise from joint interaction objectives, such as collision constraints.

\subsection{Summary}

Our approach works in two phases: 1) offline we learn a predictive model of the human $s_{t+1:T} = f(s_{0:t})$ where $s_{0:t}$ is the observed trajectory of human states. The aim of $f$ is to predict the kinematic states of the human in the next time steps up to a prediction horizon $T$, based on a sequence of previous states. This is achieved by supervised training of a position-velocity recurrent encoder-decoder neural network model on human motion capture data, 2) online we use the learned model to predict future states. The prediction is optimized to fulfill constraints by varying  the velocity inputs in the decoder's loop function at every time step.

\subsection{Human Model}
\label{ssec:human_model}

For prediction of the human we base our model on the position-velocity recurrent encoder-decoder neural network (VRED)~\cite{wang2019vred}.
We model the kinematic state of a human as a vector consisting of base position, base rotation and joint angles: $s = (p, r, j)$ which in our case is a 66 dimensional vector for full-body data. The joints are toes, ankles, knees, hips, pelvis, torso, neck, head, inner shoulders, shoulders, elbows and wrists.

As proposed in \cite{wang2019vred}, we represent angles in the exponential map representation and convert them to quaternions for the loss computation.
The velocity inputs $v$ are computed using finite differences. At each time step the positions and velocities are fed into the network, which then only predicts velocities. The new states $s'$ are obtained through a residual connection by adding the velocity to the previous state.

The full architecture can be seen in Figure~\ref{fig:architecture}. Inputs variables are depicted blue, the new predictions are depicted green.
In contrast to the original model, we stack 3 GRU cells with 1000 hidden units, which improves the prediction performance in our experiments
Additionally we do not feed the position of the base (i.e., pelvis) into the recurrent unit.
This avoids conditioning the model on world positions and leads to better generalization. We found that the same approach does not hold for the base rotation
which encodes the direction of motion.
Hence we instead offset the rotation randomly during training time to avoid overfitting.

For training, the data is sliced into same sized trajectories of 2sec.
One second is fed as input to the network and 1sec is predicted by the network.
The loss is computed on the full trajectory.
We use a squared error loss for the
base position and a quaternion loss for the base rotation and joint angles:
\begin{align}
  L &= \sum_{s_{0:T}\in\mathcal{B}} 
  \underbrace{||p'-p||^2}_\text{position} + 
  \underbrace{\sum_{t, q} \text{min}(||q'_{t} - q_t||, ||q'_t + q_t ||)}_\text{quaternion}
\end{align}
with $q_t$ being the quaternion representations.
The learning rate is set to 0.0001 and the batch size is 8.

At test time we make use of numerical optimization
to handle additional constraints.
In order to change the predictions
we add an additional connection $\delta$ to the velocity inputs
(see Figure~\ref{fig:architecture} depicted red). The network thus has additional input parameter $s'_{t+1:T} = f(s_{0:t}, \delta_{t+1:T})$. These parameters can be used to change the velocities that are predicted by the neural network and fed into the next time step which can be viewed as adding an additional acceleration term into the model.

During training $\delta$ is set to zero.
Note that the $\delta_{t}$ is fed into the network at one time step but changes the predictions for all the following time steps as well. 
In the following, the neural network $f(\delta_{t+1:T})$ will only be parameterized by $\delta$
because $s_{0:t}$ is already observed at test time and is not able to change during prediction.

\subsection{Trajectory Optimization}
\label{ssec:trajopt}

In this paper we consider optimizing predictions
with the ability to account for the following additional constraints:
\begin{itemize}
\item \textit{low-level:} human prediction should be close to the original network prediction.
\item \textit{goalset:} human prediction should end up with a hand close to a specified point.
\item \textit{collision:} human prediction should not collide with obstacles.
\item \textit{human-robot:} human prediction and a robotics agent should not collide.
\item \textit{robot smoothness:} robot trajectory should be smooth.
\item \textit{robot goal:} robot trajectory should end with an endeffector close to a specified point.
\item \textit{robot collision:} robot trajectory should not collide with obstacles.
\end{itemize}

Due the fact that $\delta$ is only used as input to the neural network and does not directly change the trajectory, it is ensured that the predicted states still ground on the network prediction. However, we add an additional loss term to fulfill the \textit{low-level} constraint:
\begin{align}
    c_c(\delta) = \|\delta\|^2
\end{align}
This term ensures that $\delta$ is not becoming too big and thus does not push the network into states that are too far from the training data to make reliable predictions.

The \textit{goal} is given as the squared distance between the human endeffector and a reference point $p^*$:
\begin{align}
c_{\text{g}}(\delta) = \|\phi_{FK}(f(\delta)_T) - p^*\|^2
\end{align}
where $\phi_{FK} : s \mapsto p$ is the forward kinematics of the human,
for example, calculating the hand position $p \in \mathbb{R}^3$. Thus 
$p_t = \phi_{FK}(f(\delta_t)))$ computes the hand position at time $t=T$,
where $T$ is the last predicted time step by the network.

We represent objects using a signed distance field (SDF).
To account for the \textit{collision} a potential function,
similar as used in CHOMP~\cite{ratliff2009chomp}, is used.
Because moving more quickly through regions with high cost
should not be penalized less, cost elements are integrated with respect to an
arc-length parameterization in the workspace.
Thus in our case, we have the following obstacle potential:
\begin{align}
 c_{\text{o}}(\delta) = \sum_{t=1}^T \exp \big\{ -\alpha~\text{SDF}(p_t) \big\} 
 \Delta_H
\end{align}
\noindent
where $  \Delta_H = \|p_{t+1}-p_t\| $, with $p_t = \phi_{FK}(f(\delta_t)))$.

Similarly we define the \textit{human-robot objective}, as a function of the human variables $\delta$ and robot states $x$ as:

\begin{align}
c_{\text{j}}(\delta, x) = \sum_{t=1}^T 
\exp \big\{ -\alpha \| p_t -x_t \|  \big\}   \Delta_H   \Delta_R
\end{align}
where $  \Delta_R = \|p^R_{t+1}-p^R_t\| $, with $p^R_t = \phi_{FK}(x_t)$.

The \textit{robot goal} $c_{Rg}$ and \textit{robot obstacle} $c_{Ro}$ are similar to the human goal and obstacle constraints with the difference that we optimize the trajectory directly and do not pass it through the neural network.

\begin{figure}[t]
\centering
\includegraphics[width=.8\columnwidth]{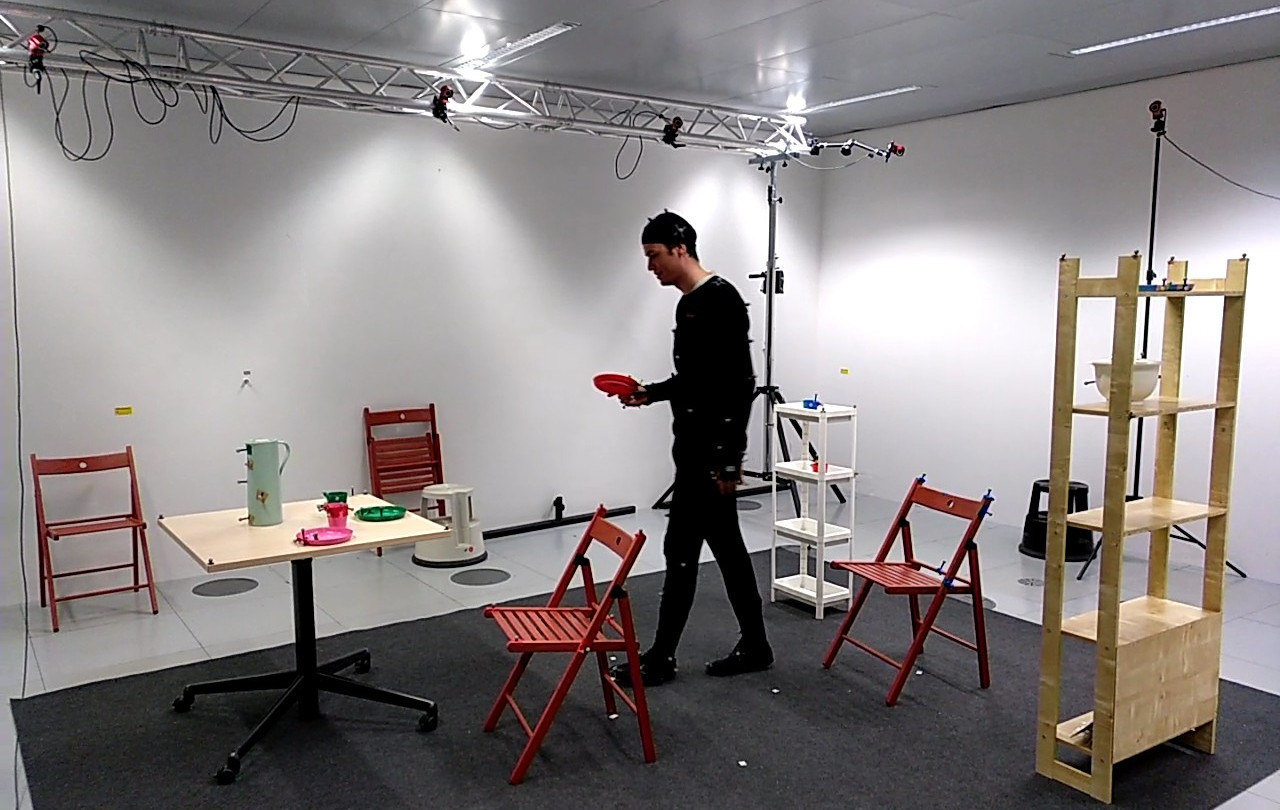}
\caption{Motion Capture Setup}
\label{fig:mocap}
\vspace{-.5cm}
\end{figure}

The \textit{robot smoothness term} is similar, as defined in CHOMP, to a sum of squared derivatives:
\begin{align}
c_{\text{smooth}}(x) = x^\top K_d x
\end{align}
where $K$ is a finite differences matrix:

\begin{align}
 K = 
 \left[
 \begin{smallmatrix}
       6  & -4	& 1  & \dots & 0 & 0 & 0 \\
       -4 & 6 & -4  & \dots & 0 & 0 & 0 \\
        1 & -4 & 6  & \dots & 0 & 0 & 0 \\
         & \vdots &  & \ddots &  & \vdots &  \\
        0 & 0 & 0  & \dots & 6 & -4 & 1 \\
        0 & 0 & 0  & \dots & -4 & 6 & -4 \\
        0 & 0 & 0  & \dots & 1 & -4 & 1 \\
  \end{smallmatrix}
  \right].
\end{align} 

The full unconstrained proxy objective is given as the sum of the cost functions for human, robot and joint cost:
\begin{align}
  c(\delta, x) &= c_H(\delta) + c_R(x) + c_J(\delta, x)\\
  c_H(\delta) &= \lambda_1 c_c(\delta) + \lambda_2 c_{\text{g}}(\delta) + \lambda_3 c_{\text{o}}(\delta)\\
  c_R(x) &= \lambda_4 c_{\text{Rg}}(x) + \lambda_5 c_{\text{Ro}}(x) + \lambda_6 c_{\text{smooth}}(x)\\
  c_J(\delta, x) &= \lambda_7 c_{\text{j}}(\delta, x)
\end{align}
with Lagrange multiplier $\lambda$.
Note that implementing this framework using a principled
augmented Lagrangian optimization algorithm would be straight forward
and is left for future work.

We derived the respective gradients and Jacobians mainly using automatic differentiation frameworks. We implemented the recurrent neural network using the Tensorflow library and we use its automatic differentiation functionality to compute the Jacobian of the network $J_{\text{NN}}=\frac{\partial f}{\partial \delta}$. We use a limited memory version of the numerical optimization algorithm BFGS in order to optimize the trajectory~\cite{byrd1995limited}.

\section{EXPERIMENTS}
\label{sec:experiments}

\begin{figure*}
\centering
\newcommand{\ltscale}{.245}
\newcommand{\vsscale}{.1cm}
\begin{subfigure}{\ltscale\textwidth}
\centering
\includegraphics[width=\linewidth]{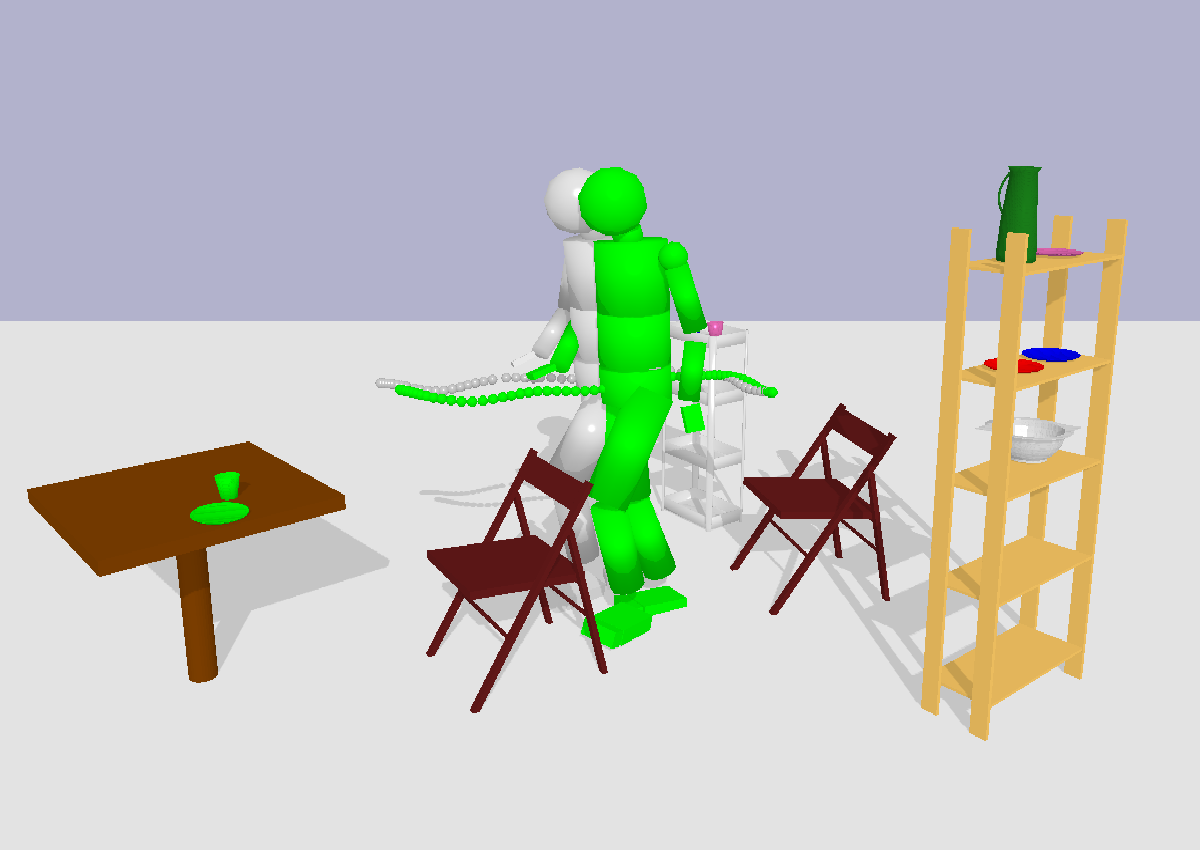}
\end{subfigure}
\begin{subfigure}{\ltscale\textwidth}
\centering
\includegraphics[width=\linewidth]{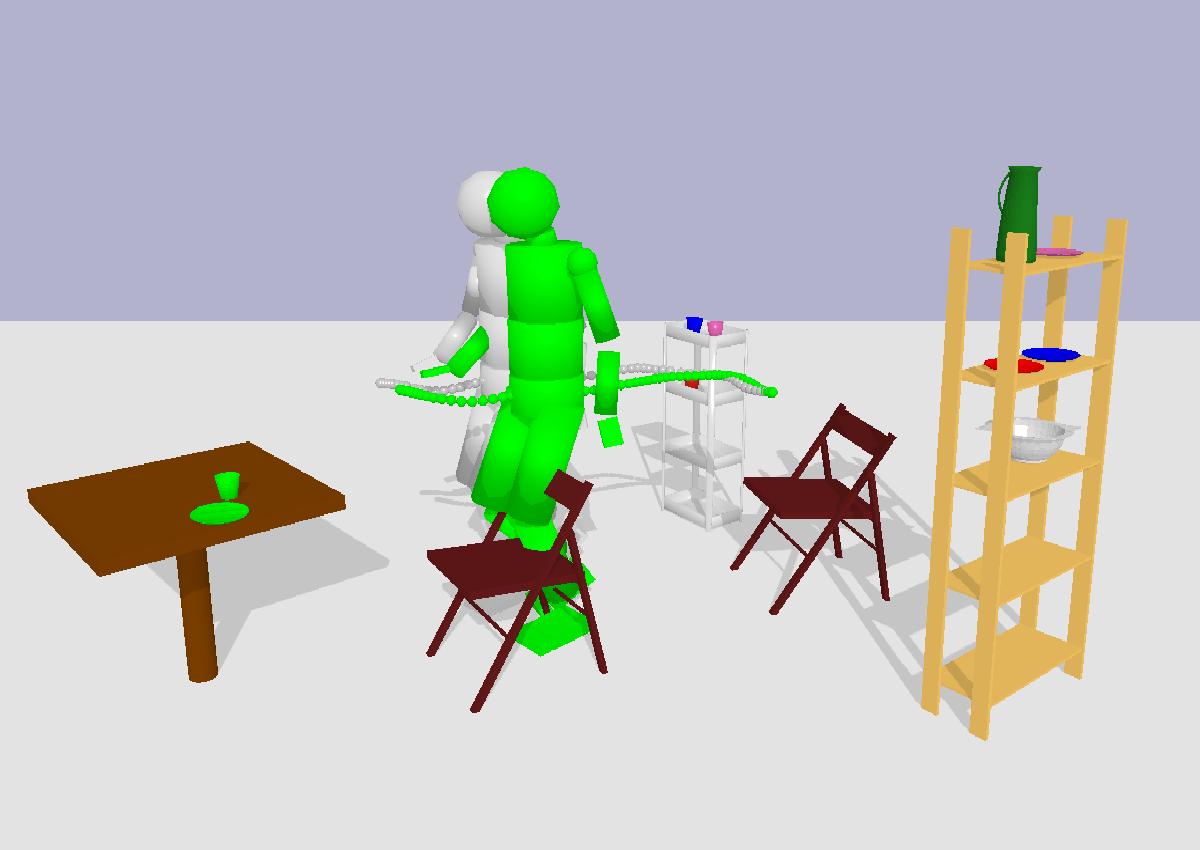}
\end{subfigure}
\begin{subfigure}{\ltscale\textwidth}
\centering
\includegraphics[width=\linewidth]{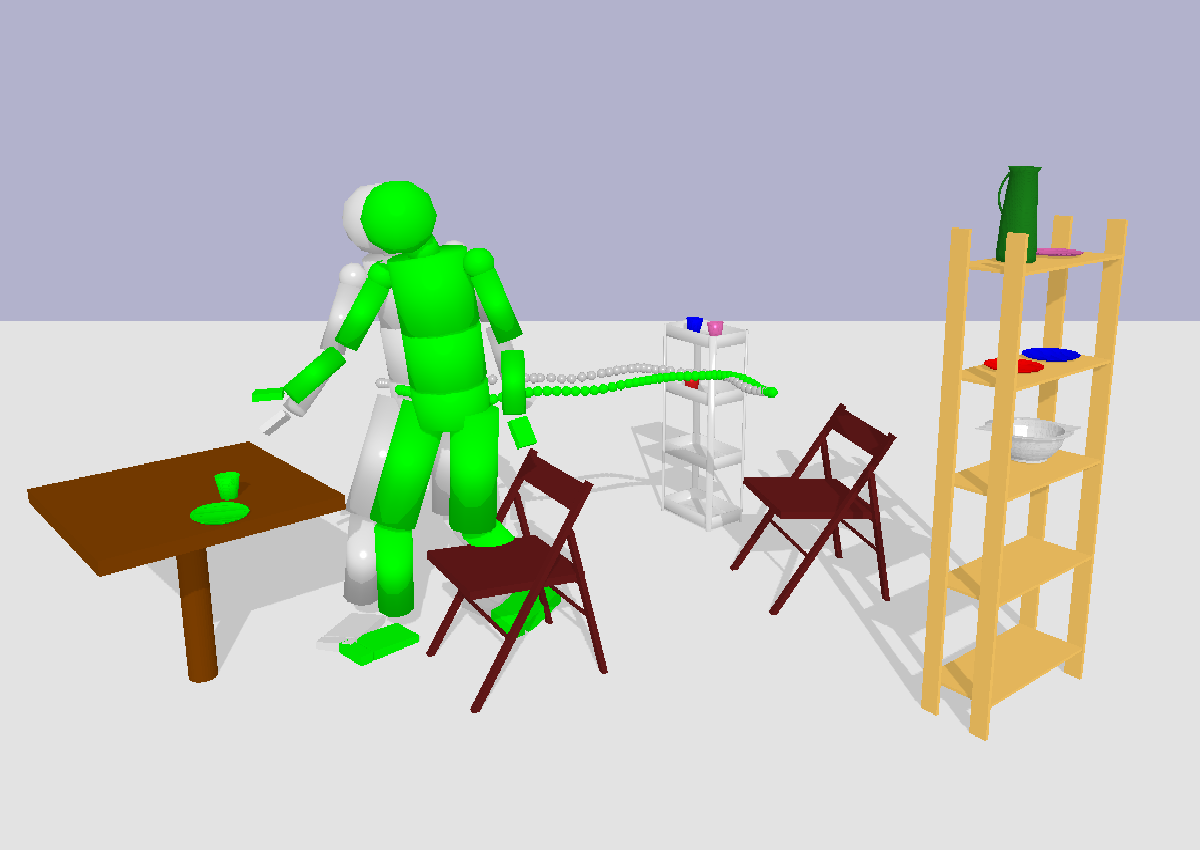}
\end{subfigure}
\begin{subfigure}{\ltscale\textwidth}
\centering
\includegraphics[width=\linewidth]{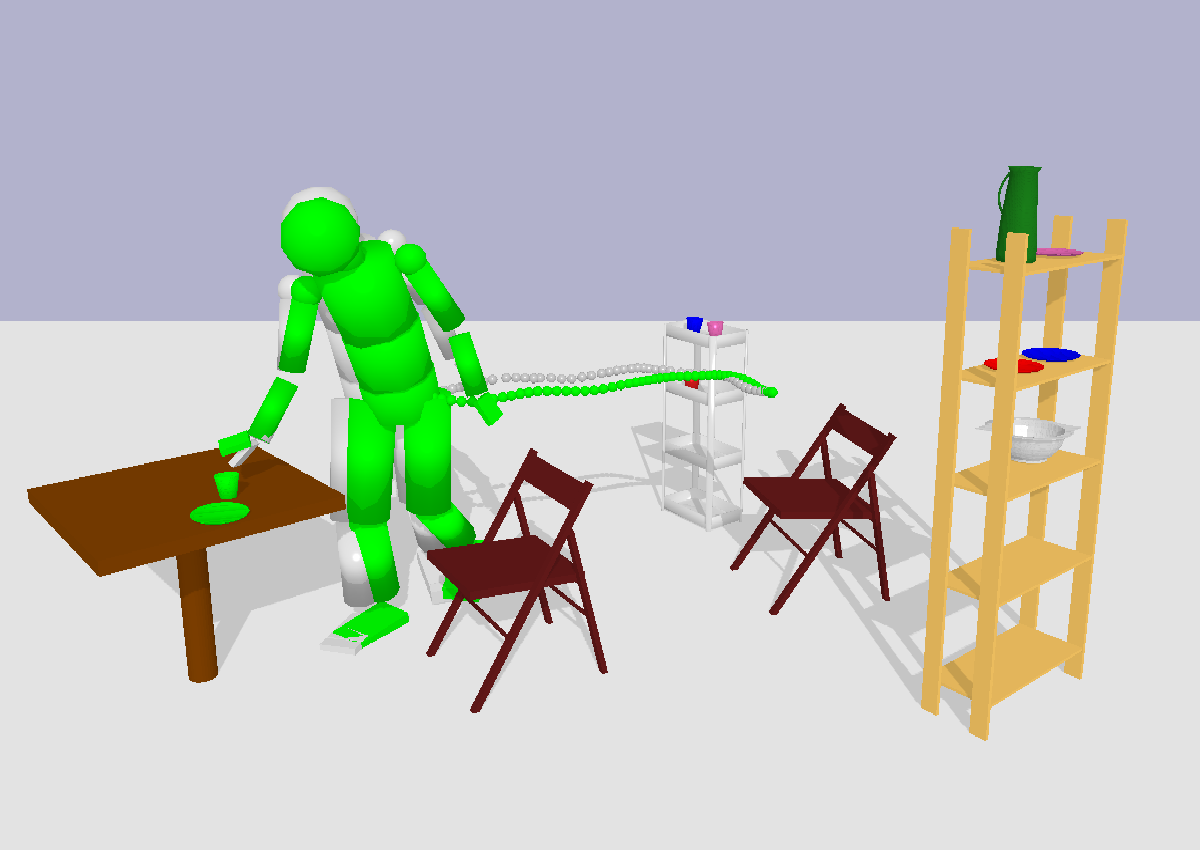}
\end{subfigure}
\\
\vspace{\vsscale}
\begin{subfigure}{\ltscale\textwidth}
\centering
\includegraphics[width=\linewidth]{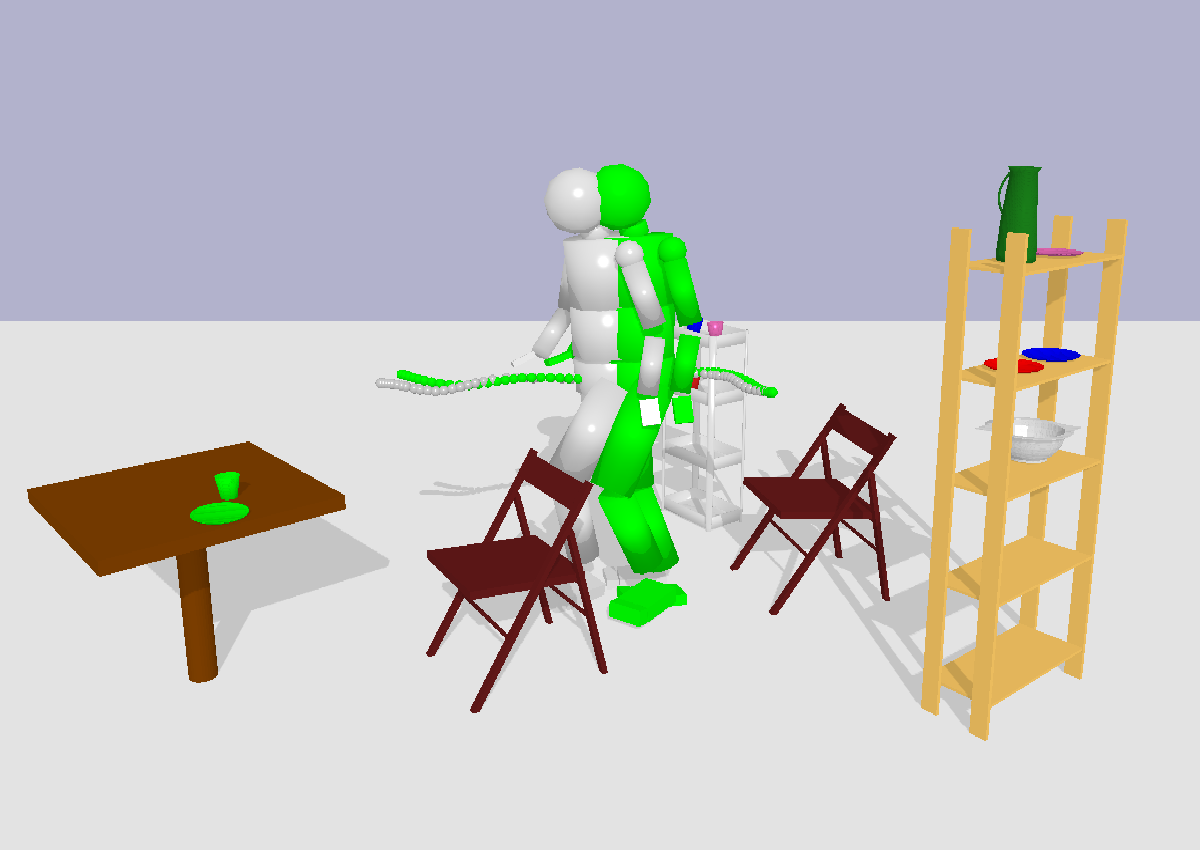}
\end{subfigure}
\begin{subfigure}{\ltscale\textwidth}
\centering
\includegraphics[width=\linewidth]{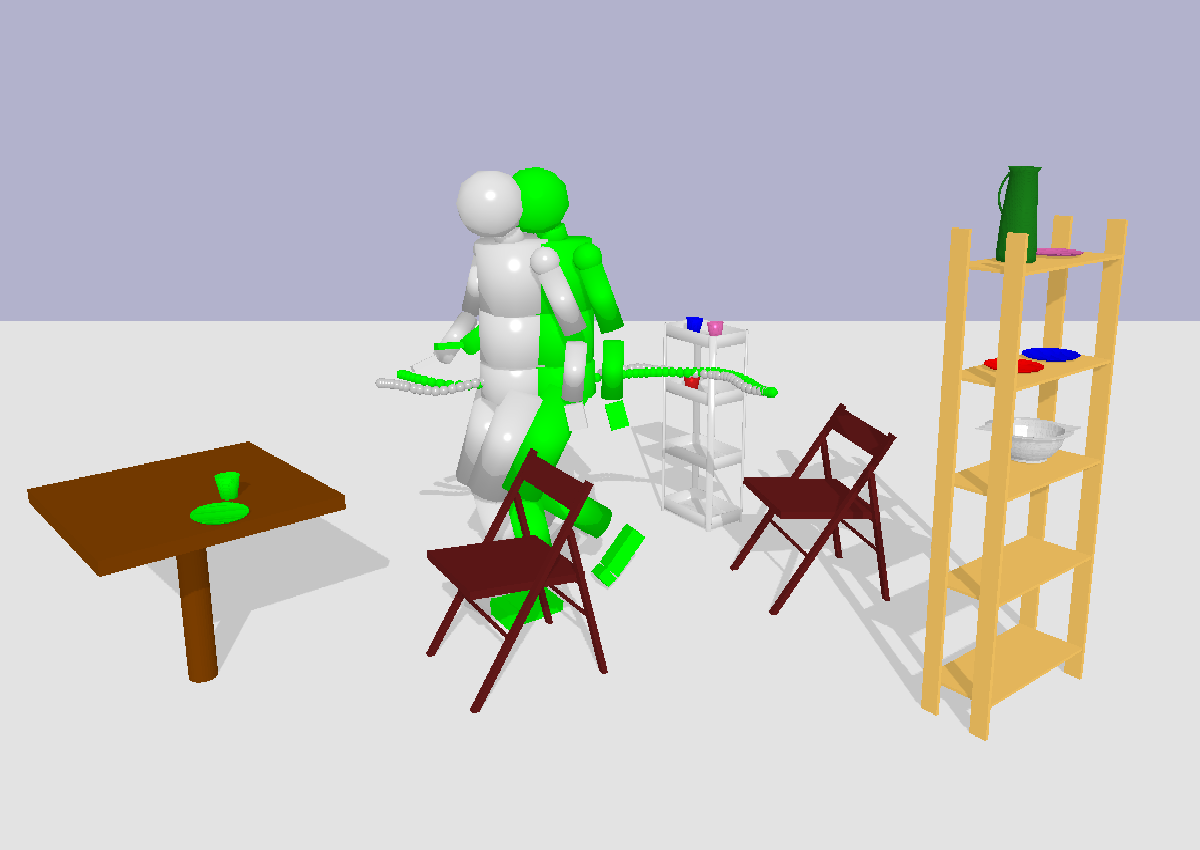}
\end{subfigure}
\begin{subfigure}{\ltscale\textwidth}
\centering
\includegraphics[width=\linewidth]{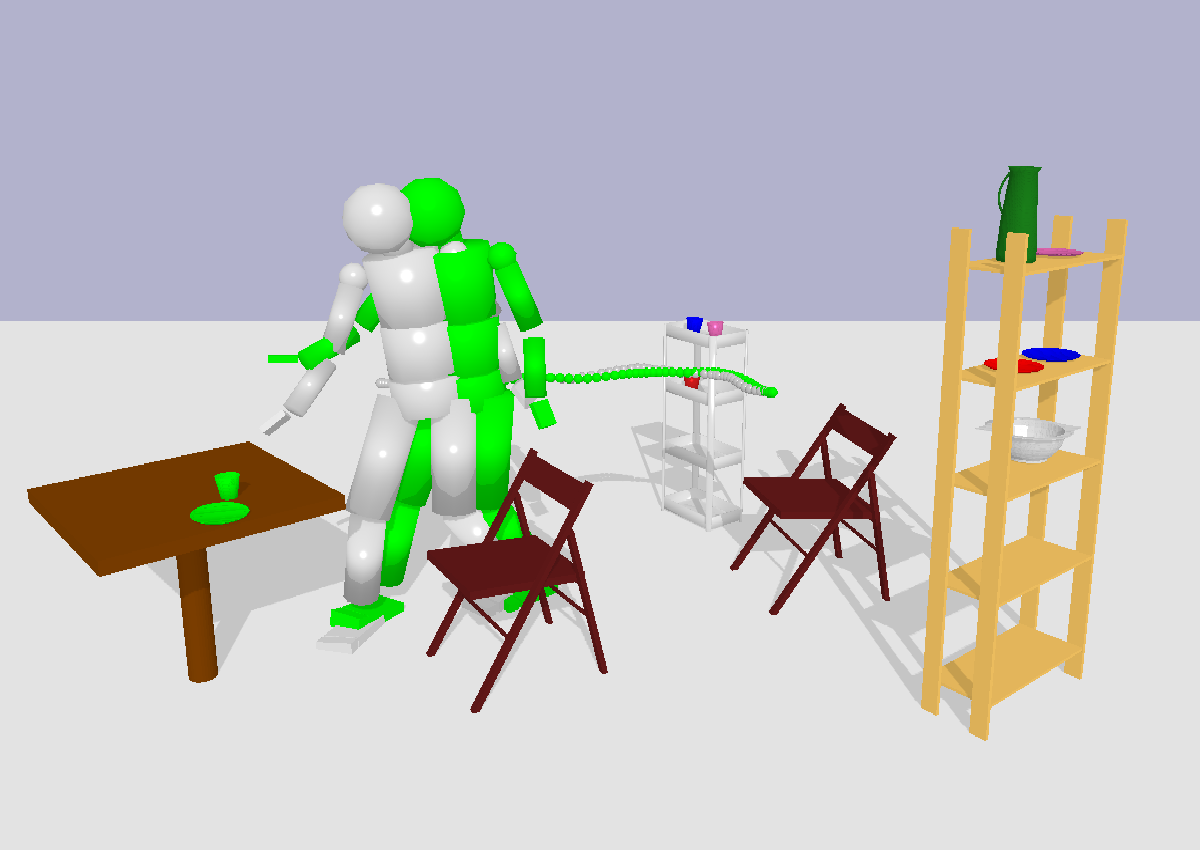}
\end{subfigure}
\begin{subfigure}{\ltscale\textwidth}
\centering
\includegraphics[width=\linewidth]{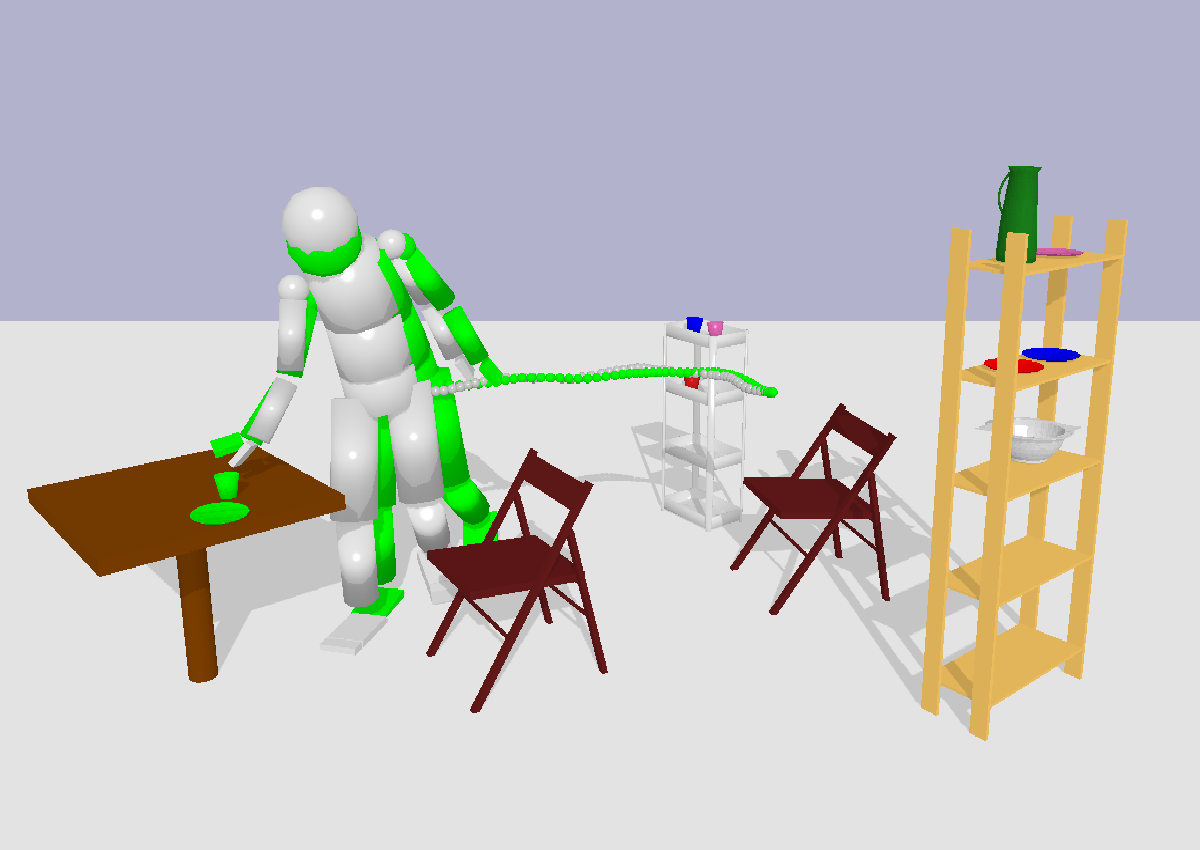}
\end{subfigure}
\caption{Human trajectory of reaching towards an object on the table for 2secs. The prediction by our method is shown in green, baseline of human motion in gray. From left to right we show the prediction after 1sec, 1.3sec, 1.6sec and 2sec. The top row shows the prediction without obstacle constraint, the bottom row with obstacle constraint. In the prediction without obstacle constraint the human collides with the chair.}
\label{fig:example_coll_traj}
  \vspace{-.5cm}
\end{figure*}

In this section we evaluate our method using real motion data.
Since available motion capture datasets do not include obstacles,
we gathered our own dataset, which can be made available on request.

All data was captured using an Optitrack motion capture system with 50 reflecting markers placed on the entire human body. The motion capture setup can be seen in Figure~\ref{fig:mocap}.  Data was recorded at a rate of 120Hz and downsampled to 30Hz for use in the network.
For the experiments we captured 3 datasets using one single actor.
A \textit{reaching1} dataset, which contains pick and place motions of different objects on different heights and lasts 31 minutes.
A \textit{walking} dataset, which consists of 60 minutes of walking data in the motion capture area. A \textit{reaching2} dataset, which contains 22 minutes of pick and place motions of different objects similar to \textit{reaching1}, however, chairs have been placed in the scene so that the human has to perform the pick and place tasks while walking around the chairs.
We train our model with data from all three datasets.
A fraction of 10\% from every dataset is held out for testing purposes.

\subsection{Goal Set Constraints}

In our first experiment we evaluate whether the prediction of the VRED network can be improved by our prediction method when we already know the target position of the reaching motion for the human hand.

From the \textit{reaching1} test set 25 reaching trajectories with a length of 1sec have been extracted. We compare our method with three baselines. We compute the distance of key joints of the human (wrists, elbows, knees, ankles and pelvis) and compute the mean distance of the predictions to the ground truth (see Table~\ref{tab:pred_methods}).

The zero velocity baseline predicts the same state for all future steps. The VRED baseline is the prediction network without the trajectory optimization part. Our method is informed with the goal position of the hand. Table~\ref{tab:pred_methods}~(b) shows the sum of the mean distances of the 9 key joints. The use of trajectory optimization improves the prediction among all future steps.

In Table~\ref{tab:pred_methods}~(w) we only compute the distance of the wrist to the ground truth. We also compute a linear interpolation baseline between the start position of the wrist and the target position. The interpolation baseline and our method are informed with the goal state and thus able to get zero error in the last time step.
However, our method, where the recurrent neural network implicitly
reconstructs the underlying human dynamics,
outperforms the interpolation baseline 
on other time steps, which simply constructs a naive straight line.

The results show that our method is able to reconstruct the full-body trajectory of the human given a target state which is useful for scenarios where we already know possible target states of the human obtained by a higher level prediction mechanism or through scene understanding, for example using affordances~\cite{koppula2016anticipating}.

A limitation of the approach with goal set constraint is that we have to predefine the duration of the trajectory which is a known problem in the trajectory optimization literature~\cite{ratliff2009chomp}. Simple approaches to solve this issues include optimizing trajectories of different time horizon or reoptimizing the trajectory with a longer horizon if the target position is not reached. A more general approach would dynamically add and remove samples during optimization. %TODO?

\begin{table}
  \centering
\setlength\tabcolsep{5.5pt}
  \begin{tabular}{r|c|c|c|c|c|c|c|c}
  ms & 125 &   250 &   375 &   500 &   625 &   750 &   875 &  1000\\
    \hline\hline
    Zerovel (b) & 0.72 & 1.37 & 1.80 & 2.31 & 2.72 & 3.01 & 3.15 & 3.25\\
    VRED (b) & \textbf{0.20} & 0.36 & 0.45 & 0.57 & 0.68 & 0.78 & 0.86 & 0.94 \\
    ours (b) & \textbf{0.20} & \textbf{0.35} & \textbf{0.44} & \textbf{0.53} & \textbf{0.56} & \textbf{0.59} & \textbf{0.62} & \textbf{0.64} \\
   \hline \hline
    Zerovel (w)  & 0.14 & 0.28 & 0.37 & 0.48 & 0.56 & 0.61 & 0.62 & 0.62 \\
    VRED (w)  & \textbf{0.03} & \textbf{0.07} & 0.09 & 0.11 & 0.13 & 0.14 & 0.15 & 0.15 \\
    ours (w) & \textbf{0.03} & \textbf{0.07} & \textbf{0.08} & \textbf{0.09} & \textbf{0.08} & \textbf{0.06} & \textbf{0.04} & 0.01 \\
    Interp (w) & 0.05 & 0.11 & 0.15 & 0.17 & 0.17 & 0.13 & 0.08 & \textbf{0.00}\\
  \end{tabular}
  \caption{Error of state prediction on different time steps in the future for the whole body~(b) and the right wrist only~(w). Reported values are in meters. For the whole body the sum distance of 9 key joints is shown.}
  \label{tab:pred_methods}
  \vspace{-.5cm}
\end{table}

\begin{figure*}
  \centering
  \newcommand{\ltscale}{.245}
\newcommand{\vsscale}{.1cm}
\vspace{\vsscale}
\begin{subfigure}{\ltscale\textwidth}
\centering
\includegraphics[width=\linewidth]{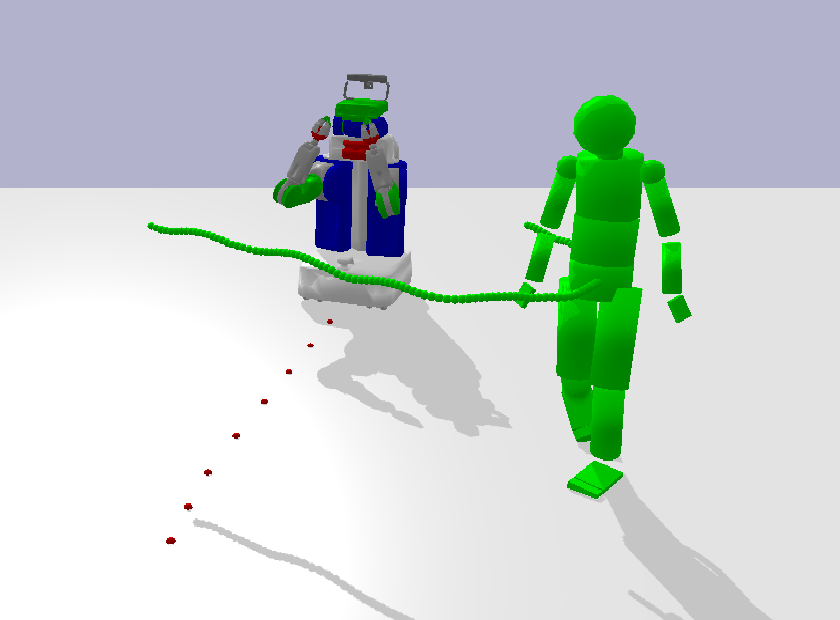}
\end{subfigure}
\begin{subfigure}{\ltscale\textwidth}
\centering
\includegraphics[width=\linewidth]{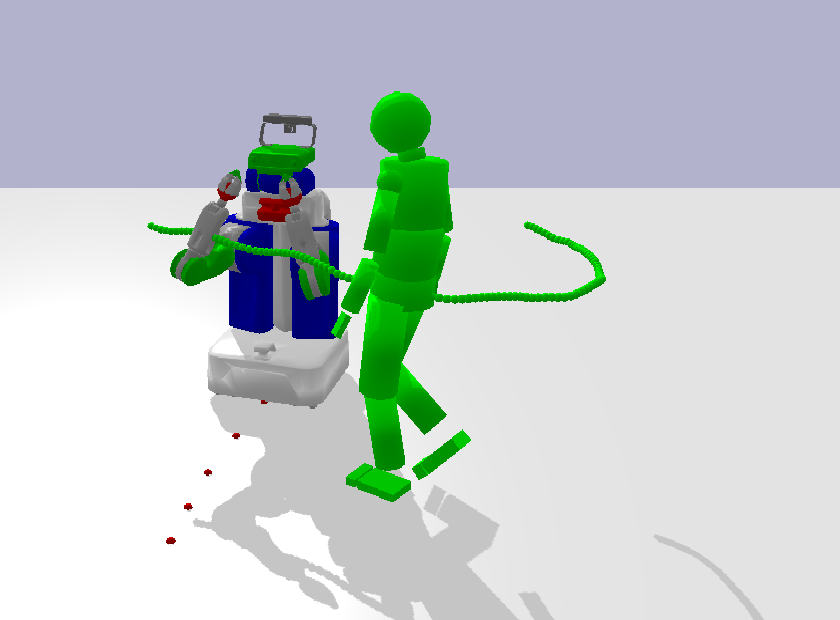}
\end{subfigure}
\begin{subfigure}{\ltscale\textwidth}
\centering
\includegraphics[width=\linewidth]{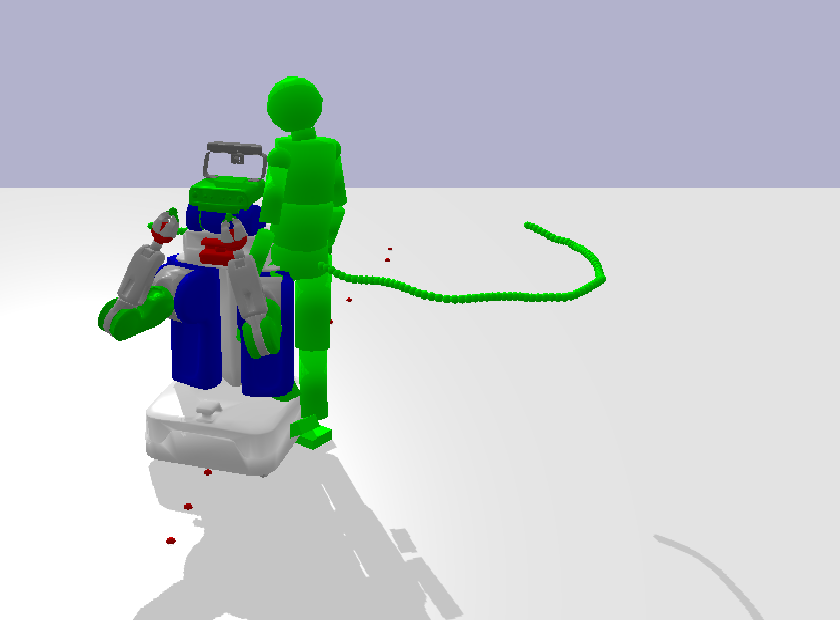}
\end{subfigure}
\begin{subfigure}{\ltscale\textwidth}
\centering
\includegraphics[width=\linewidth]{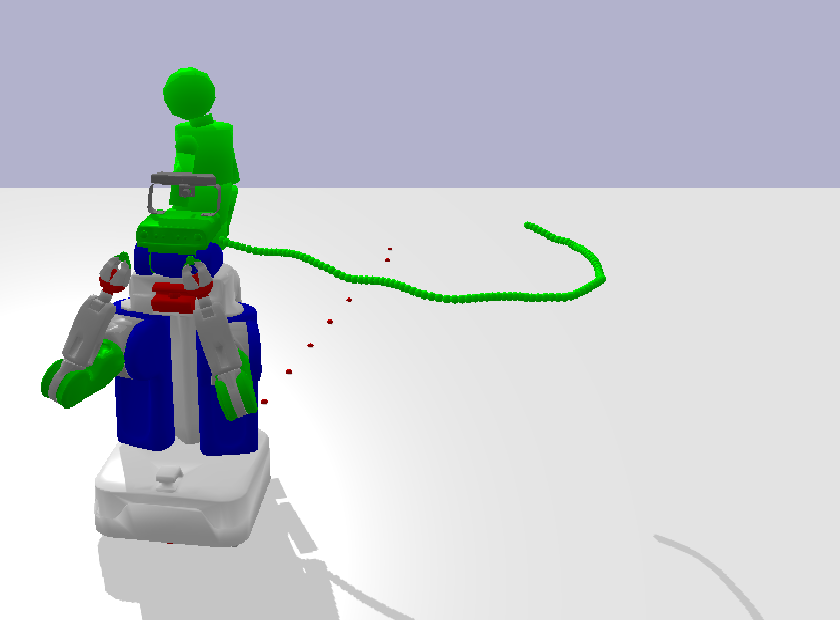}
\end{subfigure}
\\
\vspace{\vsscale}
\begin{subfigure}{\ltscale\textwidth}
\centering
\includegraphics[width=\linewidth]{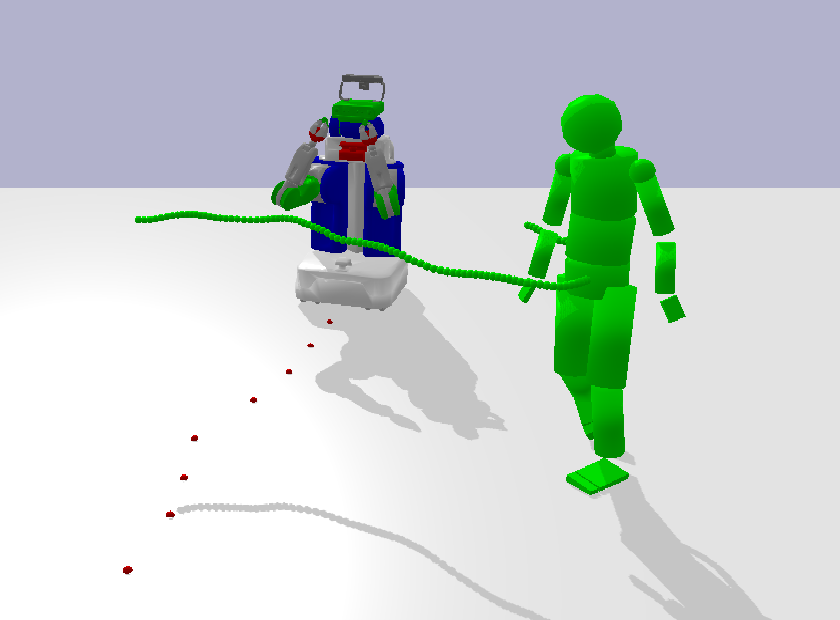}
\end{subfigure}
\begin{subfigure}{\ltscale\textwidth}
\centering
\includegraphics[width=\linewidth]{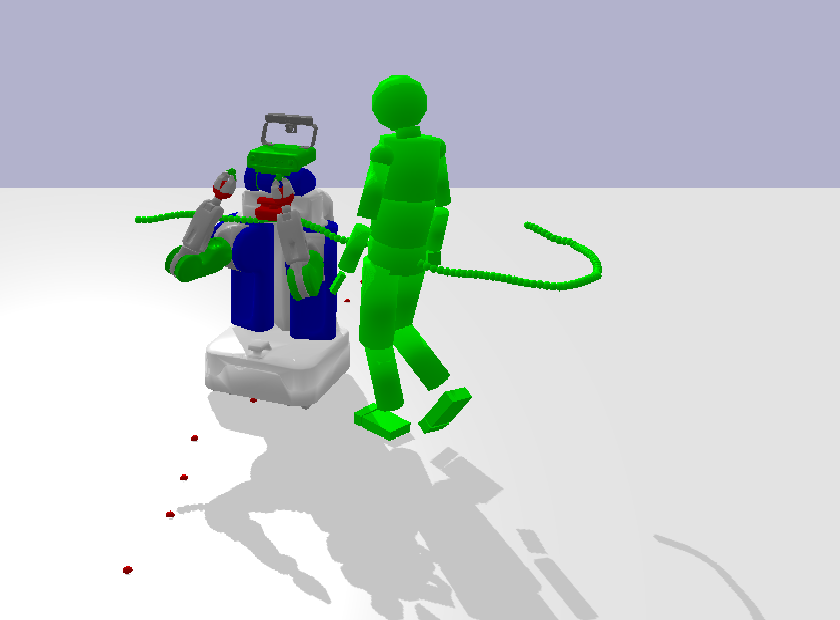}
\end{subfigure}
\begin{subfigure}{\ltscale\textwidth}
\centering
\includegraphics[width=\linewidth]{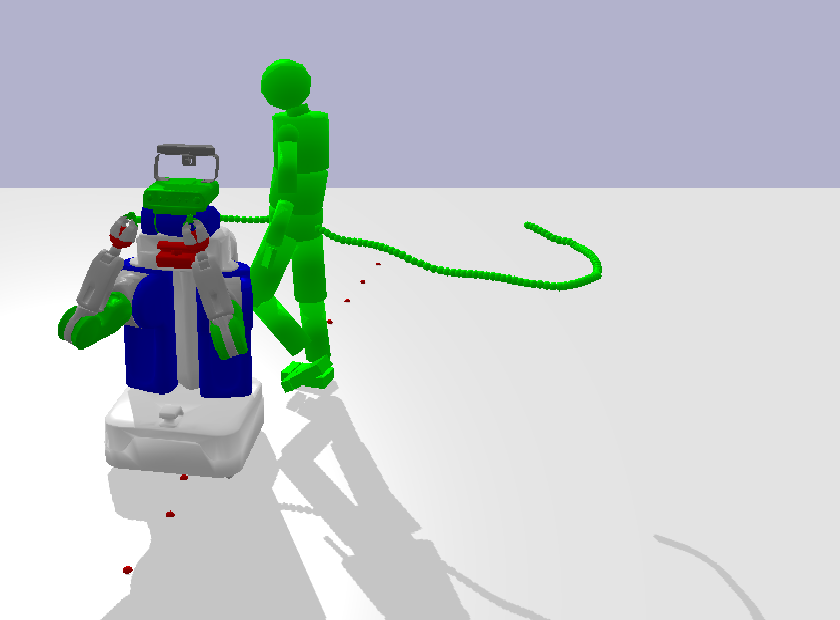}
\end{subfigure}
\begin{subfigure}{\ltscale\textwidth}
\centering
\includegraphics[width=\linewidth]{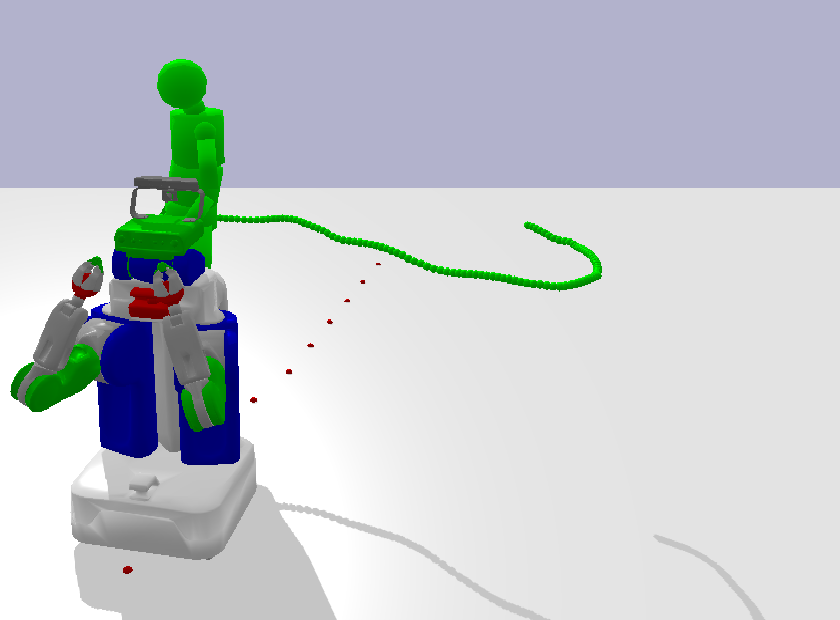}
\end{subfigure}
\caption{Joint trajectory optimization of human and robot. From left to right the trajectories after 1sec, 2.3sec, 3sec and 3.6sec. The top row shows the trajectory without the joint optimization potential, the bottom trajectory is with joint optimization.}
\label{fig:example_joint_traj}
  \vspace{-.5cm}
\end{figure*}
\subsection{Obstacle Constraints}
\begin{table}
  \centering
\setlength\tabcolsep{5.5pt}
  \begin{tabular}{r|c|c|c|c|c|c|c|c}
  ms &  250 &  500 &  750 & 1000 & 1250 & 1500 & 1750 & 2000 \\
    \hline\hline
    Zerovel & 1.21 & 2.45 & 3.90 & 5.50 & 7.53 & 9.49 & 11.15 & 11.74 \\
    VRED & 0.70 & 1.25 & 1.79 & 2.48 & 3.48 & 4.56 & 5.70 & 6.39 \\
    ours g & \textbf{0.68} & \textbf{1.25} & 1.79 & 2.27 & 2.57 & 2.52 & 2.24 & 2.18 \\
    ours g+o & 0.69 & \textbf{1.25} & \textbf{1.74} & \textbf{2.18} & \textbf{2.45} & \textbf{2.41} & \textbf{2.15} & \textbf{2.10} \\
  \end{tabular}
  \caption{Error of state prediction on different time steps in the future for the whole body. Our method with goal objective (g) and goal and obstacle objectives (g+o) is shown.}
  \label{tab:pred_methodsoc}
  \vspace{-.5cm}
\end{table}
In this experiment we evaluate whether the obstacle potential helps to further improve the prediction. Therefore we extracted 22 reaching trajectories from the \textit{reaching2} test set with a length of 2sec. Some of the trajectories contain motion trajectories that are close to an obstacle (chair). We approximate the chair sign distance field as a sphere and add an obstacle objective in addition to the goal objective.

Figure~\ref{fig:example_coll_traj} shows an example reaching trajectory with the ground truth (gray) and our prediction (green). The top row shows the prediction without obstacle objective. There the human prediction collides with the chair because the RNN has no information about obstacles in the scene. Activating the obstacle objective in the trajectory optimization step penalizes collisions and forces the prediction around the object. In the second row the improved prediction with obstacle avoidance is shown. The prediction no longer collides with the chair.  

We performed a quantitative comparison averaged over the 22 extracted reaching motions (see Table~\ref{tab:pred_methodsoc}). The table shows the mean full body error of the key joints compared to the ground truth. Note that only a few of the trajectories contain motion trajectories that are close to the obstacle. The method with additional obstacle objective outperforms the other prediction methods.

\subsection{Joint Human Robot Optimization}
The last experiment is about joint optimization of human and robot for collaborative planning. We initialize a human walking trajectory from the test data of the \textit{walking} dataset. Additionally, we add a robot model. We set a goal objective for the human base and the robot base so that the trajectories of human and robot intersect. We initialize the robot trajectory as a straight line from start to target position and the human trajectory with setting the $\delta$ to zero. The trajectories are predicted for a duration of 4sec. Using the goal objectives without the human-robot objective will lead to a collision of robot and human (see Figure~\ref{fig:example_joint_traj} top). Only planning for the robot and assuming that the human stays on the predicted trajectory would lead to a route where the robot completely avoids the human and has to go for a longer path around. In reality this is not necessary because the human would react to the robot moving and adapt his or her walking path to the robot moving. Similar, if we keep the robot trajectory fixed, the human would either need to wait till the robot passes or walk a longer way around the robot, which results in reaching the goal state later. Our aim is to find a collision free plan for human and robot so that both, the human and the robot have to leave the optimal route only a bit. Therefore we jointly optimize the robot states $x$ and the human prediction variables $\delta$ including the human-robot objective. The resulting trajectory can be seen in Figure~\ref{fig:example_joint_traj} bottom. The robot speeds up a bit and keeps farther away from the human and the human prediction walks a bit further to the right.

The experiment shows that our method can be used to plan a robot trajectory while simultaneously adapting the human prediction to it. This can be used to plan a trajectory that minimally changes the human path. How much the robot or the human has to deviate from the optimal path can be modified by adapting the optimization parameters $\lambda$.
\section{CONCLUSIONS}
\label{sec:conclusions}
In this paper we presented a novel prediction framework for human motion in the presence of environmental objectives. We showed that a state-of-the-art recurrent neural network model can be adapted to use it within a trajectory optimization framework. This improves the predictions and accounts to unseen environmental structures.

Furthermore, we showed an initial experiment on how the method could be used for shared human-robot planning.

For future work we plan to conduct real world experiments on real human-robot interaction tasks. We plan to optimize the trajectory optimization parameters in a way to adapt for human comfort during more complex interaction scenarios

\section*{ACKNOWLEDGMENT}
This work is partially funded by the research alliance ``System Mensch''.
The authors thank the International Max Planck Research School for Intelligent Systems (IMPRS-IS) for supporting Philipp Kratzer.

\bibliographystyle{IEEEtran}
\balance
\bibliography{bibliography}

\end{document}